\documentclass[runningheads]{llncs}
\usepackage[T1]{fontenc}
\usepackage{newtxtext}       % 
\usepackage[varvw]{newtxmath}       % selects Times Roman as basic font

\usepackage{graphicx}
\usepackage[hidelinks]{hyperref}
\usepackage{booktabs}
\usepackage{cleveref}
\usepackage{subcaption}
\usepackage{bm}

\usepackage[ruled,vlined]{algorithm2e} 
\SetKwInput{KwInput}{Input}
\SetKwInput{KwOutput}{Output}

\crefname{algocf}{algorithm}{algorithms}
\Crefname{algocf}{Algorithm}{Algorithms}

\begin{document}
\title{Bayesian Active Clustering with Oracle Noise}
\author{Anri Patron\inst{1}\orcidID{0000-0002-4728-0391} \and
Collin Leiber\inst{2}\orcidID{0000-0001-5368-5697} \and
Kai Puolam\"aki\inst{1}\orcidID{0000-0003-1819-1047}}
\authorrunning{A. Patron et al.}
\institute{University of Helsinki, Helsinki, Finland\\\email{\{anri.patron,kai.puolamaki\}@helsinki.fi} \and
Aalto University, Espoo, Finland\\
\email{collin.leiber@aalto.fi}
}
\maketitle              % typeset the header of the contribution
\begin{abstract}
Active clustering methods query an oracle for pairwise similarity constraints (must/cannot-links) and use the answers to steer the clustering toward the partition aligned with their domain knowledge. Existing methods assume noise-free oracle answers or a known error rate, fix the number of clusters in advance, or ignore the distance information. We propose Bayesian Active Clustering with Oracle Noise (BACON), a Bayesian active clustering method that combines a nearest-neighbor distance-dependent Chinese restaurant process prior with a noisy-oracle likelihood. The number of clusters and the oracle's error rate are inferred jointly with the partition, so erroneous answers can be overridden by later evidence. Queries are selected by expected information gain, which reduces to uncertainty sampling on posterior predictive must-link probabilities. On 16 datasets across tabular, text, and image data with error rates up to 20\%, BACON achieves the best average rank against the state of the art.

\end{abstract}

\section{Introduction}

Clustering is widely used to discover structures in unlabeled data. Methods such as $k$-means \cite{kmeans} rely solely on measures of similarity or distance for extracting the clusters. However, due to the large combinatorial space, unsupervised clustering methods might not find solutions that correspond with the user's understanding of the data. Supervised methods, on the other hand, require labeled data, which are expensive and time-consuming to collect. Consequently, semi-supervised algorithms incorporate pairwise constraints indicating binary similarity (must/cannot-links) to guide the algorithm toward the desired partition \cite{basu2004probabilistic,basu2004active}. For clustering, asking for binary pairwise similarity is usually easier to specify than labels, which require an established taxonomy.

Typically, the constraints are provided by human experts (``oracles''), which means the answers might be noisy due to ambiguous pairs or simple mistakes. As errors can propagate due to the transitivity of must-links \cite{soenen2021tackling}, they can substantially impact the final clustering. The oracle's error rate is typically not known in advance, which complicates the issue further.
Each query is costly as human effort is involved in labeling, which means we usually can only collect a small fraction of all possible pairwise constraints. This suggests treating the problem similar to active learning where the most informative pairs are selected for querying \cite{vancraenendonck2018cobras,patron2026codac}. An effective method should therefore be robust to oracle noise, select informative queries, and combine the received constraints effectively with the unsupervised information (such as distances or similarities).

{\bf Previous work.}
Previous work on semi-supervised clustering can be roughly divided into passive and active methods, depending on how the constraints are selected. Passive methods (also known as \emph{constrained clustering}) assume a set of constraints is available and try to find a partition that minimizes the constraint violations \cite{wagstaff2001constrained,basu2004probabilistic}. Active methods, referred to as \emph{active clustering} \cite{craenendonck2018cobra}, iteratively query an oracle for new constraints and refine the partition with each answer. Because of the active selection process, these methods obtain more informative constraints than a random selection and usually outperform the passive methods with the same number of constraints \cite{basu2004active,patron2026codac}.

Many methods select the queries by heuristics. 
Basu, et al. \cite{basu2004active} uses farthest-first selection maximizing diversity.
COBRAS \cite{vancraenendonck2018cobras} splits and merges prototype clusters based on proximity and size to refine the final clustering result. Its extension, NCOBRAS, asks ``redundant'' verification queries that are implied by existing constraints and probabilistically filters out the noisy constraints given a user-specified error rate \cite{soenen2021tackling}.
Early probabilistic work by Vlachos et al. \cite{vlachos2010active} uses the Chinese restaurant process with modified Gibbs sampling that forces must-linked points into the same clusters and cannot-linked points into different clusters, therefore requiring noise-free queries. They select the queries by maximizing co-assignment uncertainty.

Pei et al. \cite{pei2015bayesian} propose a Bayesian logistic regression model for active clustering, requiring both the number of clusters $k$ and the oracle error rate to be fixed a priori, which are often unknown in real scenarios. They propose two query selection strategies, maximizing uncertainty and informativeness.
Aronsson and Chehreghani \cite{aronsson2024correlation} proposed an active query selection procedure for correlation clustering (ACC) that can infer $k$ and handles noise by querying pairs multiple times. They later extended the work by proposing information-theoretic query selection strategies \cite{aronsson2025informationtheoretic}. The correlation clustering model uses only oracle-queried similarity information and disregards distance information \cite{aronsson2024correlation,aronsson2025informationtheoretic}, which can be wasteful.

\begin{figure}[htb]
    \centering
    \includegraphics[width=\linewidth]{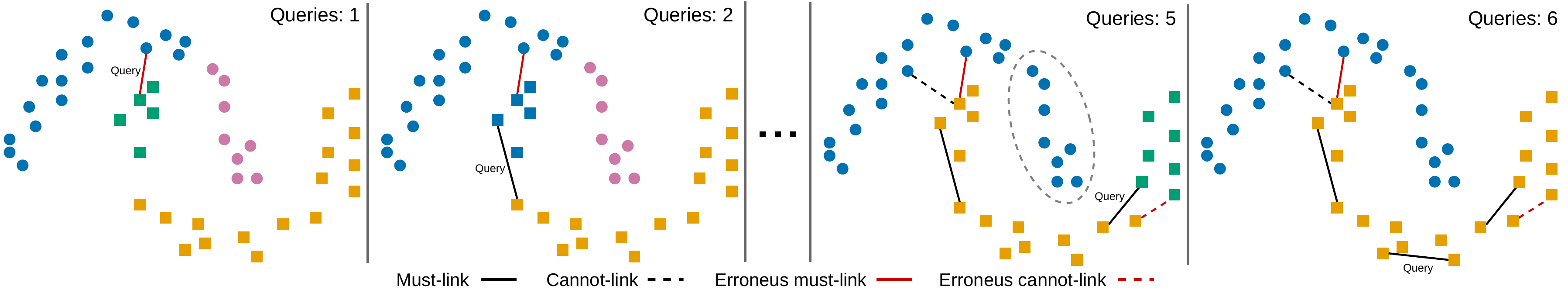}
    \caption{Visual illustration of our BACON algorithm. Shapes denote the ground truth and colors the cluster labels; the last queried pair is annotated. The leftmost panel shows the situation before adjusting the model after the first constraint (an erroneous must-link). After 6 queries (rightmost panel) the errors were corrected by new evidence. The pink cluster was merged without being queried directly due to inferred cluster scale.}
    \label{fig:example}
\end{figure}

{\bf Contributions.}
From above, only NCOBRAS infers $k$, is robust to noisy oracle answers, and exploits the unsupervised information in the data, but it needs a fixed error rate as input.
We propose \textit{Bayesian Active Clustering with Oracle Noise} (BACON)\footnote{Reproducible code available at \url{https://github.com/edahelsinki/BACON}}, a probabilistic model with all three properties. Specifically, BACON (i) integrates unsupervised information using nearest-neighbor distance-dependent Chinese restaurant process prior, under which each point tends to share a cluster with its neighbors and through it infers $k$, (ii) uses a noisy-oracle likelihood whose error rate is inferred jointly with the partition, so that erroneous answers can be overridden by later evidence, and (iii) selects queries by expected information gain, which reduces to uncertainty sampling on posterior predictive must-link probabilities. BACON is illustrated in \Cref{fig:example}.

\section{Method}

In this section, we introduce the BACON probabilistic model, describe an Markov chain Monte Carlo (MCMC) algorithm for sampling its posterior, and describe the query selection maximizing expected information gain that reduces to uncertainty selection on pair independent oracle noise. 
Lastly, we discuss the scalability of BACON to large datasets and describe a simple compression strategy for scalable active clustering.

\subsection{Problem Setting}

Let $\mathcal{X}=\{ \mathbf{x}_1,\dots, \mathbf{x}_n\}$ be a dataset with Euclidean distances $d_{ij}=\|\mathbf{x}_i-\mathbf{x}_j\|$, collected in $\mathbf{D}=(d_{ij}) \in \mathbb{R}^{n \times n}$. We represent a partition of $[n]=\{1,\dots,n\}$ using a label vector $\mathbf{c}=(c_1,\dots,c_n)$ and let $\delta_{ij}(\mathbf{c})=\mathbf{1}[c_i=c_j]$. Our goal is to recover an unknown true partition $\mathbf{c}^*$ by querying an oracle about pairs of points with a query budget $q$. For a queried pair $(i_t,j_t)$ the oracle returns $y_{t}\in\{0,1\}$ (cannot/must-link), with $\operatorname{Pr}(\delta_{i_tj_t}(\mathbf{c}^*) = y_{t}) = 1 - \epsilon$. The error rate $\epsilon \in[0,\frac12)$ is unknown and is inferred together with the partition. Each pair is queried at most once, as repeated answers from one oracle are unlikely to be independent. After $t$ queries, we denote the oracle answers with $\mathbf{y}= (y_1, \dots, y_{t})$ and the set of constraints collected so far with $\{(i_1, j_1, y_{1}), \dots , (i_t,j_t, y_{t})\} = \mathcal{Q}$.

\subsection{Generative Model and Inference}\label{sec:model}
\begin{figure}[tb]
    \centering
    \includegraphics[width=0.98\linewidth]{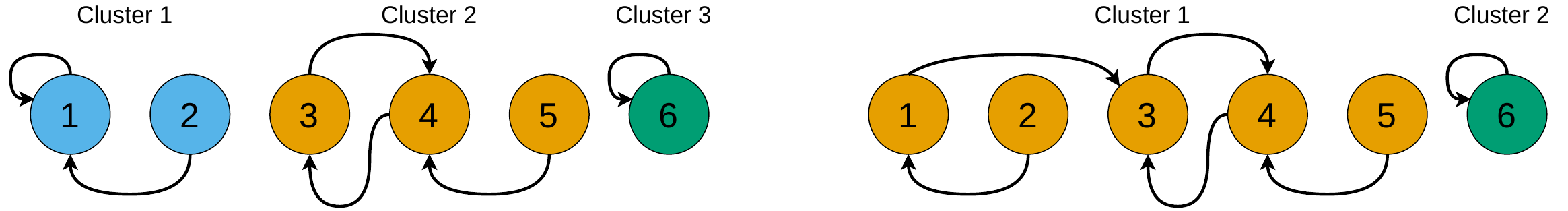}
    \caption{Visual example of how links $\mathbf{z}$ induce a clustering using two configurations (left and right). Circles denote the data points and edges the links. Here, setting $z_1 = 3$ (arrow from point 1 to point 3) merges clusters 1 and 2 (right figure).}
    \label{fig:linking_example}
\end{figure}

In this section, we describe a probabilistic clustering model that is guided by oracle provided constraints.
We allow clusters to be manifold-like, where two points may share a cluster through a chain of nearby points. We model this with a distance-dependent Chinese restaurant process (dd-CRP) \cite{blei2011distance} restricted to nearest neighbors. Each point $i$ links to exactly one point, denoted by $z_i\in\{i\}\cup \mathcal{N}_\kappa(i)$, where $\mathcal{N}_\kappa(i)$ denotes the the $\kappa$ nearest neighbors of $i$. We treat each link as an undirected edge $\{i,z_i\}$ on nodes $[n]$ in graph $G$. The clustering $\mathbf{c} = g(\mathbf{z})$, where $\mathbf{z} = (z_1, \dots, z_n)$, is given by the connected components of $G$, i.e., points $i$ and $j$ share a cluster iff they are joined by a path of links (see a visual example in \Cref{fig:linking_example}). Note that several different link configurations can induce the same clustering $\mathbf{c}$. The restriction to nearest neighbors rules out distant links and reduces the number of candidate links for each point from $n$ to $\kappa + 1$.

Following Blei and Frazier~\cite{blei2011distance}, the prior on $\mathbf{z}$ is defined as
\begin{equation}\label{eq:ddcrp}
    p(\mathbf{z} | \mathbf{D}, \alpha, \lambda^2) = \prod_{i=1}^n p(z_i| \mathbf{D}, \alpha, \lambda^2) = \prod_{i=1}^n \Bigg[ \frac{ {\bf 1}[z_i = i] \alpha +  {\bf 1}[z_i \neq i]  f(d_{i z_i}, \lambda^2)}{\alpha + \sum_{j \in \mathcal{N}_\kappa(i)} f(d_{ij}, \lambda^2)} \Bigg],
\end{equation}
where $f(d_{ij}, \lambda^2) = \exp(- d_{ij}^2 / 2\lambda^2)$ is a decay function, $\alpha \geq 0$ the self-link concentration, and $\lambda^2  > 0$ the length scale parameter. So if $z_i = i$ then $p(z_i| \mathbf{D}, \alpha, \lambda^2) \propto \alpha$ and otherwise $p(z_i| \mathbf{D}, \alpha, \lambda^2) \propto f(d_{i z_i}, \lambda^2)$. On the parameters, we place priors $\alpha \sim \text{Gamma}(a_1, b_1)$ and $\lambda^2 \sim \text{Inv-Gamma}(a_2, b_2)$.
Because of the nearest neighbor restriction, the points in the same cluster must be connected in the $\kappa$-nearest-neighbor graph; clusters with points in disconnected components receive zero prior probability and are never sampled.

We denote the number of constraints in $\mathcal{Q}$ that $\mathbf{c}$ violates with $m(\mathbf{c})=\sum_{t=1}^{|\mathcal{Q}|} \mathbf{1}[y_t\neq\delta_{i_tj_t}(\mathbf{c})]$. 
The constraints $\mathcal{Q}$ are included in the model via the noisy-oracle likelihood
\begin{equation}\label{eq:oracle}
    p(\mathbf{y} | \mathbf{c}, \epsilon) = \prod_{t=1}^{|\mathcal{Q}|} p(y_t \mid i_t, j_t, \mathbf{c}, \epsilon) = \epsilon^{m(\mathbf{c})} (1 - \epsilon)^{|\mathcal{Q}| - m(\mathbf{c})}.
\end{equation}
We treat $\epsilon$ as a latent variable and place on it a Beta prior truncated to $[0, \tfrac12)$, i.e., $p(\epsilon) \propto \epsilon^{a_3 - 1} (1-\epsilon)^{b_3 - 1} \mathbf{1}[0 \leq \epsilon < \tfrac12]$.
Limiting $\epsilon < \tfrac12$ prevents the model from interpreting the oracle's answers as more wrong than right. 
The full joint posterior is
\begin{equation}\label{eq:posterior}
    p(\mathbf{z}, \alpha, \lambda^2, \epsilon | \mathbf{D}, \mathcal{Q}) \propto p(\mathbf{z} | \mathbf{D}, \alpha, \lambda^2) p(\mathbf{y}| \mathbf{c}=g(\mathbf{z}), \epsilon) p(\alpha) p(\lambda^2) p(\epsilon).
\end{equation}

For sampling \Cref{eq:posterior}, we use an MCMC algorithm, described in \Cref{alg:mcmc}. Each link $z_i$ is drawn from its full conditional over its $\kappa+1$ candidates (Gibbs sampling) as in Blei and Frazier~\cite{blei2011distance}, $\epsilon$ is updated by a Gibbs step, and $\alpha$ and $\lambda^2$ are updated by Metropolis-Hastings with log-normal proposals (details given below). To extract a partition, i.e., compute $\mathbf{c} = g(\mathbf{z})$, we use the union-find algorithm \cite{cormen2009introduction}.

{\bf Sampling $\alpha$, $\lambda^2$ and $\epsilon$.}
Next, we provide the conditional distributions needed in \Cref{alg:mcmc}. 
We denote the number of self links with $u = \sum_{i=1}^n {\bf 1}[z_i = i]$ and let $\omega_{i} = \alpha + \sum_{j \in \mathcal{N}_\kappa(i)} f(d_{ij}, \lambda^2)$.
First, we treat the concentration parameter $\alpha$. The Gamma density (shape-scale) is $p(\alpha) \propto \alpha ^{a_1-1} \exp(-\alpha/b_1)$. The conditional is
\begin{align}
p(\alpha | \mathbf{z}, \lambda^2, \mathbf{D}) \propto \alpha ^{u+a_1-1} \exp(-\alpha/b_1) \prod_{i=1}^n \omega_{i}^{-1}.
\end{align}
Next, we will consider the length-scale parameter $\lambda^2$.
The inverse-Gamma density is $p(\lambda^2) \propto (\lambda^2)^{-(a_2 + 1)} \exp(-b_2 / \lambda^2)$.
The conditional density is
\begin{align}
p(\lambda^2 | \mathbf{z}, \alpha, \mathbf{D}) \propto (\lambda^2)^{-(a_2 + 1)} \exp \Bigg(\frac{-b_2 - \frac{1}{2} \sum_{i:z_i \neq i} d_{i z_i}^2}{\lambda^2}\Bigg) \prod_{i=1}^n \omega_{i}^{-1} .
\end{align}
Both distributions are non-standard, but Metropolis-Hastings (MH) algorithm can be used to sample from them.
Lastly, the conditional distribution for the error rate $\epsilon$ is
\begin{align}
p(\epsilon | \mathbf{c}=g(\mathbf{z}), \mathcal{Q}) \propto \epsilon^{a_3 + m(\mathbf{c}) -1} (1-\epsilon)^{b_3 + |\mathcal{Q}| - m(\mathbf{c}) - 1} \mathbf{1}[0 \leq \epsilon < \tfrac{1}{2}],
\end{align}
which is a truncated $\text{Beta}(a_3 + m(\mathbf{c}), b_3 + |\mathcal{Q}| - m(\mathbf{c}))$ distribution. As the number of violations $m(\mathbf{c})$ grows, the probability mass shifts towards $\tfrac{1}{2}$ and large $\epsilon$ values decrease the penalty for partitions violating constraints.

\begin{algorithm}[tb]
\DontPrintSemicolon
\caption{MCMC sampler for BACON posterior}
\label{alg:mcmc}
\KwInput{Initial values $\mathbf{z},\alpha, \lambda^2, \epsilon$, distances $\mathbf{D}$, constraints $\mathcal{Q}$, hyperparameters $\kappa, a_1, b_1, a_2, b_2, a_3, b_3$, number of samples $S$, burn-in samples $S_{0}$}
\For{$s \in [S + S_0]$}{
    \For{$i \in [n]$}{
        \For{$l \in \mathcal{N}_{\kappa}(i) \cup \{i\}$}{
            $p_l \gets p(z_i = l \mid \mathbf{D}, \alpha, \lambda^2)\, p(\mathbf{y} \mid g(\mathbf{z}_{-i}, z_i = l), \epsilon)$ \tcp{Unnormalized conditional, see \Cref{eq:ddcrp,eq:oracle}}
        }
        Sample $z_i = l$ with probability $\propto p_l$ \tcp{Gibbs step}
    }
    $\alpha \gets \operatorname{MH\_sample}(\alpha, \mathbf{z}, \lambda^2, \mathbf{D}, a_1, b_1) \quad \lambda^2 \gets \operatorname{MH\_sample}(\lambda^2, \mathbf{z}, \alpha, \mathbf{D}, a_2, b_2)$\;
    $\mathbf{c} \gets g(\mathbf{z}) \quad \epsilon \gets \operatorname{sample\_truncated\_Beta}(a_3 + m(\mathbf{c}), b_3 + |\mathcal{Q}|-m(\mathbf{c}), 0, \tfrac{1}{2})$\;
    \If{$s > S_{0}$}{ 
        $\mathbf{c}^{(s - S_0)} \gets \mathbf{c}$, \quad $\epsilon^{(s - S_0)} \gets \epsilon$\;
    }
}
\KwOutput{Samples $\mathcal{S} = ((\mathbf{c}^{(s)}, \epsilon^{(s)}))_{s=1}^{S}$ and final state $\mathbf{z}, \alpha, \lambda^2, \epsilon$}
\end{algorithm}

{\bf Marginal posterior.} Query selection and the final clustering depend on $\mathbf{z}$ only through the partition $\mathbf{c} = g(\mathbf{z})$, and so $\alpha$ and $\lambda^2$ can be marginalized. The marginal $p(\mathbf{c}, \epsilon | \mathbf{D}, \mathcal{Q}) = \sum_{\mathbf{z} : g(\mathbf{z}) = \mathbf{c}} \int\!\!\int p(\mathbf{z}, \alpha, \lambda^2, \epsilon | \mathbf{D}, \mathcal{Q}) \, d\alpha \, d\lambda^2$ is never computed explicitly, since the pairs $(\mathbf{c}^{(s)}, \epsilon^{(s)})$ from \Cref{alg:mcmc} are draws from the marginal.
By also marginalizing over the error rate $\epsilon$ gives $p(\mathbf{c}| \mathbf{D}, \mathcal{Q})$. The same-cluster probability of points $i,j$ and its estimate are
\begin{equation}
    \rho_{ij} = \operatorname{Pr}(c_i = c_j | \mathbf{D}, \mathcal{Q}) = \mathbb{E}_{\mathbf{c} | \mathbf{D}, \mathcal{Q}}[\delta_{ij}(\mathbf{c})] \qquad \widehat{\rho}_{ij} = \frac{1}{S}\sum_{s=1}^S \delta_{ij}(\mathbf{c}^{(s)}).
\end{equation}

{\bf Point estimation.}
To get hard clustering point estimates, we use the Binder loss \cite{wade2018bayesian}, for a candidate partition $\mathbf{c}'$ and the true partition $\mathbf{c}^*$, it is defined as
\begin{equation} \label{eq:binder_loss}
    L(\mathbf{c}', \mathbf{c}^*) = \sum_{i<j} {\bf 1}[c'_i = c'_j] {\bf 1}[c^*_i \neq c^*_j] + {\bf 1}[c'_i \neq c'_j] {\bf 1}[c^*_i = c^*_j].
\end{equation}
Since $\mathbf{c}^*$ is unknown, we pick the partition that minimizes the loss \Cref{eq:binder_loss} in expectation over the posterior among the MCMC samples $\mathbf{c}^{(1)}, \dots, \mathbf{c}^{(S)}$ \cite{dahl2006modelbased,wade2018bayesian}, i.e., $\hat{\mathbf{c}} = \operatorname{argmin}_\mathbf{c'} \mathbb{E}_{\mathbf{c} \sim p(\mathbf{c} | \mathbf{D}, \mathcal{Q})}[L(\mathbf{c'}, \mathbf{c})] 
= \operatorname{argmin}_\mathbf{c'} \sum_{i<j} \big[\delta_{ij}(\mathbf{c'})(1-\widehat{\rho}_{ij}) + (1-\delta_{ij}(\mathbf{c'}))\widehat{\rho}_{ij}\big]$.

{\bf Computational complexity.} For one sample \Cref{alg:mcmc} loops over $n$ points each with $\kappa +1$ link candidates. For each candidate, extracting the labels with union-find takes takes $O(n)$ time up to inverse-Ackermann factor \cite{cormen2009introduction}, and evaluating the oracle likelihood takes $O(|\mathcal{Q}|)$ time. Finally, we multiply by the total number of samples, so the time complexity is $O(n (\kappa +1) (n + |\mathcal{Q}|) (S + S_0))$.

\subsection{Posterior-based query selection} \label{sec:query_selection}

We select the point pair whose oracle answer is expected to reduce the posterior uncertainty about the partition the most. For a candidate pair $(i,j) \in \mathcal{U}$, where $\mathcal{U}$ is the set of unqueried pairs, with unknown oracle answer $y_*$, the expected information gain (EIG) is the expected reduction in posterior entropy \cite{houlsby2011bayesian}:
\begin{equation}\label{eq:eig_intractable}
    \text{EIG}(i,j) = H(\mathbf{c} | \mathbf{D}, \mathcal{Q}) - \mathbb{E}_{y_* | \mathbf{D}, \mathcal{Q}}[H(\mathbf{c} | y_*, \mathbf{D}, \mathcal{Q})].
\end{equation}
Since $\mathbf{c}$ ranges over $B_n$ (Bell number of size $n$) partitions, \Cref{eq:eig_intractable} is intractable. We can see that  \Cref{eq:eig_intractable} is the conditional mutual information $I(\mathbf{c};y_* | \mathbf{D}, \mathcal{Q})$, and by symmetry it can be expressed in terms of the entropy of the binary answer
\begin{equation}\label{eq:eig}
    \text{EIG}(i,j) = H(y_* |\mathbf{D}, \mathcal{Q}) -  \mathbb{E}_{\mathbf{c} |\mathbf{D}, \mathcal{Q}} [H(y_* | \mathbf{c}, \mathbf{D}, \mathcal{Q})].
\end{equation}

Given a partition $\mathbf{c}$ and an error rate $\epsilon$, the oracle answers must-link ($y_* = 1$) with probability $(1-\epsilon)\delta_{ij}(\mathbf{c}) + \epsilon(1-\delta_{ij}(\mathbf{c})) = \epsilon + (1-2\epsilon)\delta_{ij}(\mathbf{c})$. Marginalizing over the joint posterior of $(\mathbf{c}, \epsilon)$ gives the predictive must-link probability
\begin{equation}\label{eq:pred}
    p_{ij} = \mathbb{E}_{(\mathbf{c},\epsilon) | \mathbf{D}, \mathcal{Q}}\big[\epsilon + (1-2\epsilon)\delta_{ij}(\mathbf{c})\big],
\end{equation}
so the first term of \Cref{eq:eig} is $H(y_* |\mathbf{D}, \mathcal{Q}) = H_b(p_{ij})$, where $H_b(p) = -p\log_2 p - (1-p)\log_2(1-p)$ is the binary entropy. For the second term, we fix $\mathbf{c}$ and average $\epsilon$ over its conditional (see \Cref{sec:model}), whose mean we denote by $\bar\epsilon(\mathbf{c}) = \mathbb{E}[\epsilon | \mathbf{c}, \mathcal{Q}]$. The answer then equals $\delta_{ij}(\mathbf{c})$ with probability $1 - \bar\epsilon(\mathbf{c})$, and by the symmetry $H_b(p) = H_b(1-p)$ its entropy is $H_b(\bar\epsilon(\mathbf{c}))$, which depends on $\mathbf{c}$, but not on $(i,j)$. Thus EIG is
\begin{equation}\label{eq:eig_closed}
    \text{EIG}(i,j) = H_b(p_{ij}) - \mathbb{E}_{\mathbf{c} | \mathbf{D}, \mathcal{Q}}\big[H_b(\bar\epsilon(\mathbf{c}))\big].
\end{equation}

The second term does not depend on the pair $(i,j)$, i.e., is constant, and since $H_b$ is strictly decreasing in $|p - \tfrac12|$, maximizing EIG is \emph{uncertainty sampling} on the predictive must-link probability. That is, we query the pair whose oracle answer we are least able to predict.
\Cref{eq:eig_closed} is a special case of Bayesian Active Learning by Disagreement (BALD) \cite{houlsby2011bayesian} in which the expected conditional entropy is the same for every candidate due to the assumption that the error rate $\epsilon$ is independent of the pair. In practice, however, human oracles might make more mistakes on ambiguous pairs. Using a pair-specific error rate $\epsilon_{ij}$, the EIG would favor pairs that have uncertain co-assignment but which the oracle can answer reliably \cite{houlsby2011bayesian}; for a fixed $\epsilon_{ij}$, EIG equals $H_b(\epsilon_{ij} + (1-2\epsilon_{ij})\rho_{ij}) - H_b(\epsilon_{ij})$, which is zero when $\rho_{ij} \in \{0, 1\}$ or $\epsilon_{ij} = \tfrac12$.

{\bf Selection.} Given $S$ posterior samples $(\mathbf{c}^{(s)}, \epsilon^{(s)})$, we estimate \Cref{eq:pred} by
\begin{equation}\label{eq:pred_mc}
    \widehat{p}_{ij} = \frac{1}{S}\sum_{s=1}^S \big[\epsilon^{(s)} + (1-2\epsilon^{(s)})\,\delta_{ij}(\mathbf{c}^{(s)})\big]
\end{equation}
and query $(i^t, j^t) = \operatorname{arg} \min_{(i,j) \in \mathcal{U}} |\widehat{p}_{ij} - \tfrac12|$, breaking ties randomly.
After we receive the answer $y^t$, we add the new constraint to $\mathcal{Q} \gets \mathcal{Q} \cup \{(i^t, j^t, y^t)\}$, remove the pair from $\mathcal{U} \gets \mathcal{U} \setminus \{(i^t,j^t)\}$, and sample the posterior again. The selection strategy is greedy as it always picks the most informative pair. 
We present BACON in \Cref{alg:bac}.

\begin{algorithm}[tb]
\DontPrintSemicolon
\caption{BACON}
\label{alg:bac}
\KwInput{Distances $\mathbf{D}$, query budget $q$, number of neighbors $\kappa$, hyperparameters, $S$, $S_0$}
$\mathcal{Q} \gets \emptyset, \mathcal{U} \gets \{ (i,j): 1 \le i < j \le n\}$\;
Initialize $\mathbf{z}$, $\alpha$, $\lambda^2$, $\epsilon$\;
$\mathcal{S}, \mathbf{z}, \alpha, \lambda^2, \epsilon \gets \operatorname{sample\_posterior}(\mathbf{z}, \alpha, \lambda^2, \epsilon, \mathbf{D}, \mathcal{Q})$ \tcp{see \Cref{alg:mcmc}}
\For{$t = 1, \dots, q$}{
$\widehat{p} \gets$ \Cref{eq:pred_mc} from $\mathcal{S}$ \tcp{Posterior predictive for must-link}
$(i_t, j_t)\gets \operatorname{arg} \min_{(i,j) \in \mathcal{U}} |\widehat{p}_{ij} - \tfrac12|$; query the oracle for $y_t$\; 
$\mathcal{Q} \gets \mathcal{Q} \cup\{(i_t, j_t, y_t)\}, \quad \mathcal{U} \gets \mathcal{U} \setminus \{(i_t, j_t)\}$\;
$\mathcal{S}, \mathbf{z}, \alpha, \lambda^2, \epsilon \gets \operatorname{sample\_posterior}(\mathbf{z}, \alpha, \lambda^2, \epsilon, \mathbf{D}, \mathcal{Q})$\;
}
\KwOutput{Binder point estimate $\hat{\mathbf{c}}$ from $\mathcal{S}$, samples $\mathcal{S}$}
\end{algorithm}

\begin{table}[tb]
\scriptsize
    \centering
    \caption{Datasets used in the experiments. We denote ground truth cluster count by $k_{gt}$. 
    }
    \begin{tabular}{lcccc @{\hspace{2em}} lcccc}
        \toprule
         Dataset & $n$ & $k_{gt}$ & $d$ & Domain & Dataset & $n$ & $k_{gt}$ & $d$ & Domain \\
         \midrule
         Moons & $100$ & $2$ & $2$ & Tabular & Segmentation & $2310$ & $7$ & $19$ & Tabular \\
         Iris & $150$ & $3$ & $4$ & Tabular & Waveform & $5000$ & $3$ & $21$ & Tabular \\
         Wine & $178$ & $3$ & $13$ & Tabular & BBCNews & $2225$ & $5$ & $2000$ & Text \\
         Dermatology & $358$ & $6$ & $34$ & Tabular & BBCSports & $737$ & $5$ & $2000$ & Text \\
         Ecoli & $336$ & $8$ & $7$ & Tabular & Reuters & $8367$ & $5$ & $2000$ & Text \\
         Glass & $214$ & $6$ & $9$ & Tabular & Optdigits & $5620$ & $10$ & $64$ & Image \\
         Parkinsons & $195$ & $2$ & $22$ & Tabular & USPS & $9298$ & $10$ & $256$ & Image \\
         BreastCancer & $569$ & $2$ & $30$ & Tabular & MNIST & $70\,000$ & $10$ & $784$ & Image \\
         \bottomrule
    \end{tabular}
    \label{tab:datasets}
\end{table}

{\bf Scaling to large datasets.}
For a given $\epsilon$, each oracle answer yields at most $C_{\epsilon} = 1 - H_b(\epsilon)$ bits about $\mathbf{c}$ (see \Cref{eq:eig_closed}). So, higher $\epsilon$ lowers the informativeness of the queries. By using the chain rule of mutual information, we know that $t$ arbitrarily chosen queries reduce the posterior uncertainty at most by $tC_{\epsilon}$ bits, i.e., $\mathbb{E}[H(\mathbf{c} \mid \mathbf{D}, \mathcal{Q})] \geq H(\mathbf{c} \mid \mathbf{D}) - tC_{\epsilon}$.
Without the geometric information, $H(\mathbf{c} \mid \mathbf{D})$ could be as large as $\log_2 B_n = \Theta(n \log n)$ \cite{grunwald2025explicit}. Because there are at most $(\kappa+1)^n$ link configurations for $\mathbf{z}$, and since $\mathbf{c}=g(\mathbf{z})$, it follows that $H(\mathbf{c} \mid \mathbf{D}) \leq H(\mathbf{z} \mid \mathbf{D}) \leq n \log_2(\kappa+1)$. The decay function makes the link probabilities non-uniform, favoring the closest neighbors, so the actual entropy is typically well below the bound. The entropy can still grow linearly with $n$ and so can the number of queries that are needed to resolve the partition.

For example, for MNIST ($n=70\,000$, $\kappa=5$) the bound $n\log_2(\kappa+1)$ is about $1.8\times 10^5$ bits, whereas each answer yields at most one bit, so without further assumptions the queries needed to resolve the partition may exceed practical budgets.
We, therefore, need a structural assumption that removes most of this uncertainty. To this end, we employ compression to super-instances, that are groups of points assumed to share the same cluster \cite{craenendonck2018cobra,vancraenendonck2018cobras,soenen2021tackling}.
The super-instances are found by we pre-clustering the data with $k$-means into $\tilde{n} < n$ clusters \cite{craenendonck2018cobra} and let $\sigma : [n] \to [\tilde{n}]$ be a mapping from a point to a super-instance. We represent each super-instance by the point closest to its centroid $\bm{\mu}_j$, i.e., for $j \in [\tilde{n}]$ we assign $i^*_j = \operatorname{arg\,min}_{i: \sigma(i)=j} \|\mathbf{x}_i - \bm{\mu}_j\|^2$. BACON (\Cref{alg:bac}) is then run on the $\tilde{n}$ representatives, and every point inherits the label of its super-instance $c_i = \tilde{c}_{\sigma(i)}$, where $\tilde{\mathbf{c}} = (\tilde{c}_1, \dots, \tilde{c}_{\tilde{n}})$ is the partition of the super-instances. This reduces the bound to $\tilde{n}\log_2(\kappa+1) \approx 129$ bits for $\tilde{n}=50$, at the cost of bias when super-instances are impure.
The compression also reduces the cost of the sampler as $n$ is replaced by $\tilde{n}$ in the complexity given in \Cref{sec:model}.
Making the errors in the initial compression recoverable by adding a split mechanism similar to \cite{vancraenendonck2018cobras} is left as future work.

\section{Results}

\begin{figure}[tp]
    \centering
    \begin{subfigure}{0.329\textwidth}
        \centering
        \includegraphics[width=\linewidth]{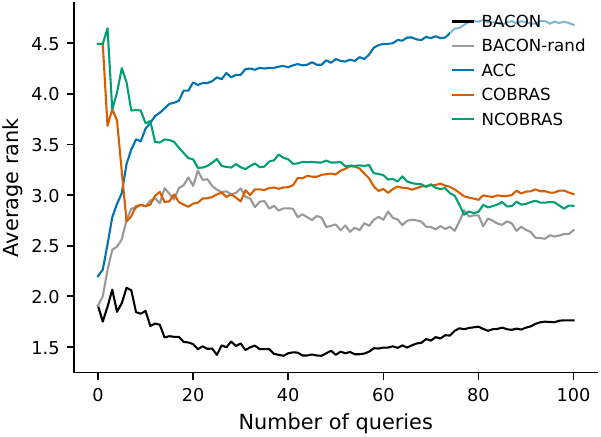}
        \caption{$\epsilon=0$}
        \label{fig:rank_eps0}
    \end{subfigure}
    \hfill
    \begin{subfigure}{0.329\textwidth}
        \centering
        \includegraphics[width=\linewidth]{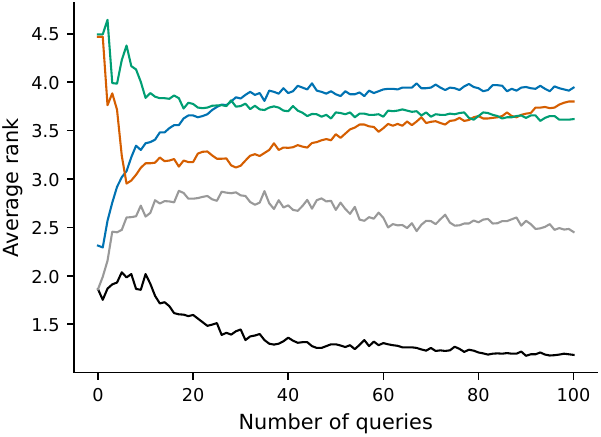}
        \caption{$\epsilon=0.1$}
        \label{fig:rank_eps01}
    \end{subfigure}
    \hfill
    \begin{subfigure}{0.329\textwidth}
        \centering
        \includegraphics[width=\linewidth]{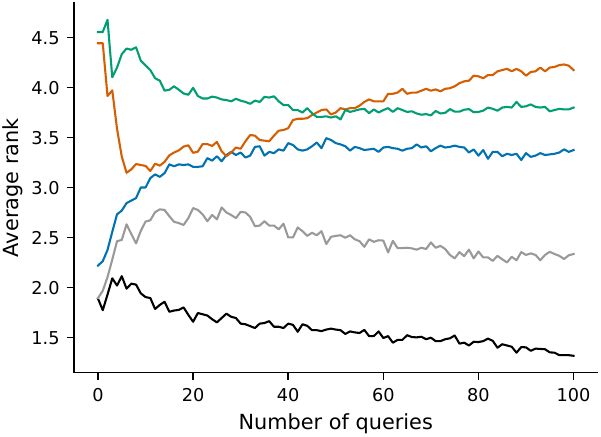}
        \caption{$\epsilon=0.2$}
        \label{fig:rank_eps02}
    \end{subfigure}
    \caption{Average ranking across datasets and repeats (lower is better).}
    \label{fig:ranking}
\end{figure}

\begin{figure}[tb]
    \centering
    \begin{subfigure}{0.38\linewidth}
        \centering
        \includegraphics[width=\linewidth]{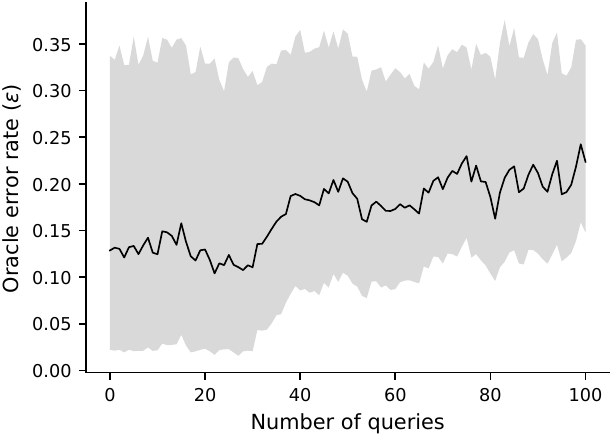}
        \caption{Inferred oracle error rate $\epsilon$}
        \label{fig:infer_eps}
    \end{subfigure}
    \hfill
    \begin{subfigure}{0.38\linewidth}
        \centering
        \includegraphics[width=\linewidth]{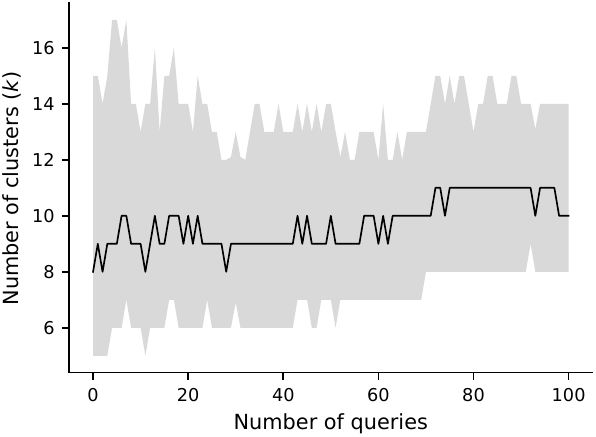}
        \caption{Inferred number of clusters $k$}
        \label{fig:infer_k}
    \end{subfigure}
    \caption{BACON posterior samples for $\epsilon$ (a) and $k$ (b) against the number of queries on MNIST (with $k_{gt}=10$ and $\epsilon=0.2$); there were $19$ oracle errors. The line denotes the median over the posterior samples and the background the $10 \%$ -- $90\%$ quantiles.}
    \label{fig:infering}
\end{figure}

In this section, we evaluate BACON against NCOBRAS and ACC that are robust to oracle noise, as well as COBRAS (a baseline assuming noise-free constraints), and a variant of BACON querying pairs randomly (BACON-rand). We use various noise levels $\epsilon$ to test their robustness. We also report ablation studies for the compression strategy (see \Cref{sec:query_selection}), the nearest neighbors $\kappa$, and oracle noise level.

{\bf Experimental setting.}
We use 16 datasets including tabular, text, and image data for evaluating the methods (see details in \Cref{tab:datasets}).
Dermatology, Glass, Parkinsons, Ecoli, Optdigits, Reuters, Segmentation, and Waveform are from the UCI repository\footnote{\url{https://archive.ics.uci.edu/}}, Moons, BreastCancer, Iris, and Wine from Scikit-learn\footnote{\url{https://scikit-learn.org/stable/api/sklearn.datasets.html}},
MNIST and USPS from Pytorch Vision\footnote{\url{https://docs.pytorch.org/vision/}}, and BBCNews and BBCSports from\footnote{\url{http://mlg.ucd.ie/datasets/bbc.html}}.
We scale the tabular data to zero mean and unit variance and use the 2000 most common term frequency–inverse document frequency (TF-IDF) features for text data. Image and text datasets are embedded to 10 dimensions by an autoencoder, where the encoder has structure $[d, 1000, 500, 500, 10]$ and the decoder mirrors the encoder; the same embedding is used by all methods.

Unless stated otherwise, we use $\kappa = 5$, $\tilde{n}=50$, $S=2000$, $S_{0}=200$ and we set the parameters for Gamma as $a_1=1, b_1=\tfrac{1}{2}$, the inverse-Gamma as $a_2=2, b_2=1$, and Beta as $a_3 = 1, b_3 = 5$. For sampling $\alpha$ and $\lambda^2$ we use Metropolis-Hastings with log-normal proposals with a step size of $1$. Finally, we scale all squared distances $d_{ij}^2$ by their median. We use the \emph{maxexp} acquisition function in ACC, as it performed the best \cite{aronsson2024correlation}, initialize the estimated pairwise similarities with $k$-means using $10$ clusters, and use the same compression strategy as with BACON ($\tilde{n}=50$) to make the algorithm scalable. NCOBRAS is given the true $\epsilon$.
We use the ACC, COBRAS, and NCOBRAS implementations from the authors.

{\bf Main results.}
We evaluate each method 10 times per dataset to control for stochasticity. Each method is ranked by the Adjusted Rand Index (ARI) \cite{hubert1985comparing} (tied ranks are averaged), and we compute the mean rank over the datasets and repeats. We present the mean ranks for oracle noise levels $\epsilon \in \{0, 0.1, 0.2\}$ in \Cref{fig:ranking} and the mean ARI after $100$ queries for $\epsilon=0.1$ in \Cref{tab:final}. NCOBRAS crashed on 0--3 runs for many datasets with $\epsilon > 0$ due to non-trivial errors; they occurred more frequently with higher $\epsilon$. The crashed runs were given the last rank and were ignored for computing \Cref{tab:final} values, and counted as zero ARI in its t-tests.

\begin{figure}[tb]
    \centering
    % \hfill
    \begin{subfigure}{0.329\linewidth}
        \centering
        \includegraphics[width=\linewidth]{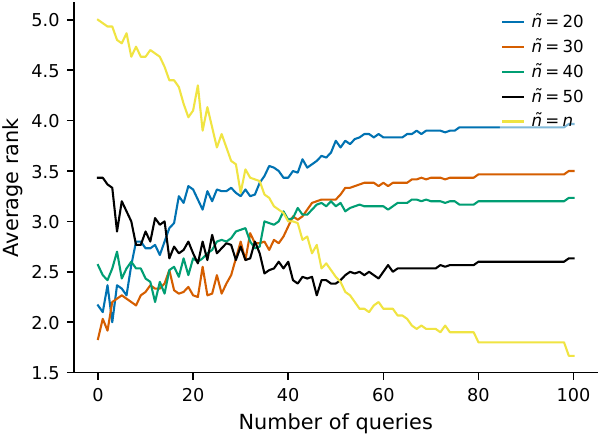}
        \caption{Compression level.}
        \label{fig:abl_compression}
    \end{subfigure}
    \hfill
    \begin{subfigure}{0.329\linewidth}
        \centering
        \includegraphics[width=\linewidth]{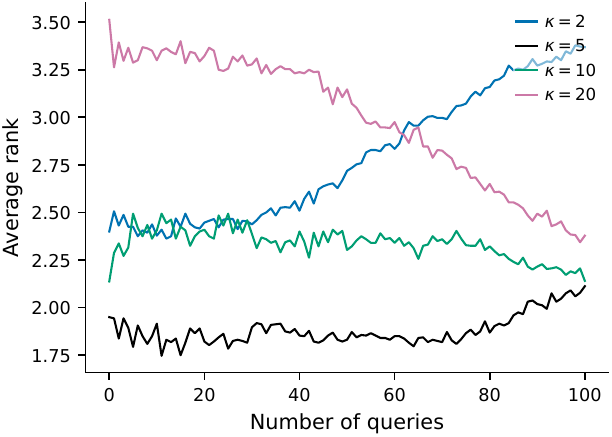}
        \caption{Nearest neighbors $\kappa$.}
        \label{fig:abl_knn}
    \end{subfigure}
    \hfill
    \begin{subfigure}{0.329\linewidth}
        \centering
        \includegraphics[width=\linewidth]{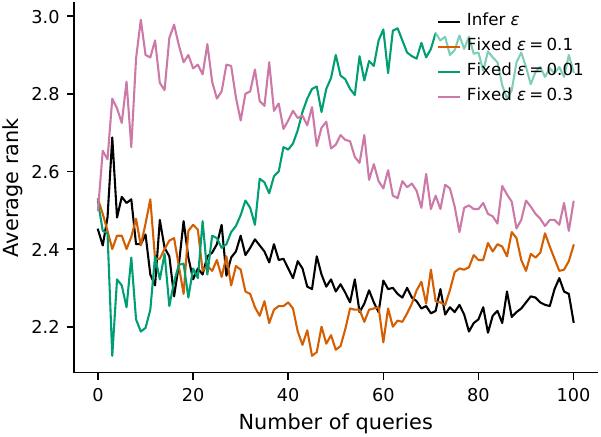}
        \caption{Oracle noise.}
        \label{fig:abl_noise}
    \end{subfigure}
    \caption{Ablation and sensitivity study results. (a) compression level ($\tilde{n}=n$ denotes no compression), (b) number of nearest neighbors $\kappa$, and (c) oracle noise: BACON inferring $\epsilon$ and BACON using fixed $\epsilon$ values $0.01$, $0.1$, and $0.3$ (with true oracle error rate $\epsilon = 0.1$). Lines denote the average rank (lower is better), see \Cref{fig:ranking}.}
    \label{fig:knn}
\end{figure}

\begin{table}[tb]
\caption{Mean ARI after $100$ constraints (with oracle error rate $\epsilon = 0.1$) and the standard deviations in parentheses. We show the largest score in {\bf bold} and \underline{underline} the scores that are not statistically lower (by one-sided t-test using significance level $0.05$). The rightmost column (Bound) shows the best ARI attainable for compressed data using $\tilde{n}=50$ and BACON\textsubscript{$q=0$} shows the performance without constraints.}
\label{tab:final}
\scriptsize
\centering
\begin{tabular}{lcccccc|c}
\toprule
 & BACON\textsubscript{$q=0$} & BACON & BACON-rand & ACC & COBRAS & NCOBRAS & Bound \\
\midrule
Moons & 0.24 (0.02) & {\bf 1.00 (0.00)} & \underline{0.96 (0.07)} & 0.28 (0.06) & 0.66 (0.13) & 0.93 (0.04) & 1.00 (0.00) \\
Iris & 0.50 (0.09) & {\bf 0.91 (0.01)} & 0.83 (0.04) & 0.46 (0.04) & 0.68 (0.11) & 0.74 (0.09) & 0.92 (0.01) \\
Wine & 0.64 (0.10) & {\bf 0.93 (0.02)} & \underline{0.92 (0.03)} & 0.61 (0.07) & 0.69 (0.15) & 0.81 (0.15) & 0.93 (0.03) \\
Dermatology & 0.74 (0.07) & {\bf 0.90 (0.02)} & 0.84 (0.06) & 0.65 (0.05) & 0.49 (0.12) & 0.73 (0.16) & 0.92 (0.02) \\
Ecoli & 0.64 (0.12) & {\bf 0.76 (0.03)} & 0.72 (0.06) & 0.63 (0.11) & 0.44 (0.16) & 0.12 (0.25) & 0.80 (0.02) \\
Glass & 0.21 (0.04) & {\bf 0.32 (0.09)} & 0.18 (0.04) & 0.22 (0.04) & 0.25 (0.07) & 0.23 (0.04) & 0.51 (0.02) \\
Parkinsons & 0.04 (0.02) & {\bf 0.71 (0.05)} & 0.62 (0.9) & 0.12 (0.06) & 0.41 (0.19) & 0.48 (0.12) & 0.71 (0.05) \\
BreastCancer & 0.39 (0.11) & {\bf 0.81 (0.04)} & \underline{0.80 (0.04)} & 0.51 (0.12) & 0.45 (0.16) & 0.67 (0.07) & 0.81 (0.03) \\
Segmentation & 0.43 (0.03) & {\bf 0.65 (0.05)} & 0.49 (0.3) & 0.37 (0.12) & 0.41 (0.08) & 0.37 (0.07) & 0.72 (0.02) \\
Waveform & 0.23 (0.03) & {\bf 0.38 (0.05)} & 0.33 (0.07) & 0.22 (0.03) & 0.32 (0.05) & \underline{0.35 (0.04)} & 0.47 (0.01) \\
BBCNews & 0.29 (0.07) & {\bf 0.74 (0.03)} & 0.57 (0.11) & 0.46 (0.05) & 0.46 (0.15) & 0.54 (0.09) & 0.75 (0.02) \\
BBCSports & 0.20 (0.07) & {\bf 0.74 (0.03)} & 0.61 (0.11) & 0.35 (0.08) & 0.40 (0.11) & 0.52 (0.11) & 0.75 (0.03) \\
Reuters & 0.44 (0.10) & {\bf 0.82 (0.03)} & 0.76 (0.08) & 0.51 (0.08) & 0.50 (0.15) & 0.54 (0.22) & 0.84 (0.01) \\
Optdigits & 0.59 (0.05) & {\bf 0.85 (0.02)} & 0.67 (0.04) & 0.70 (0.04) & 0.49 (0.14) & 0.21 (0.24) & 0.87 (0.01) \\
USPS & 0.57 (0.05) & {\bf 0.82 (0.02)} & 0.62 (0.05) & 0.61 (0.07) & 0.36 (0.15) & 0.08 (0.18) & 0.84 (0.01) \\
MNIST & 0.56 (0.06) & {\bf 0.83 (0.01)} & 0.62 (0.10) & 0.69 (0.05) & 0.45 (0.07) & 0.11 (0.21) & 0.84 (0.01) \\
\bottomrule
\end{tabular}
\end{table}

\begin{table}[tb]
\scriptsize
\centering
\caption{Average runtime over 10 repeats (in seconds) with number of queries $q=100$ for BACON using $\tilde{n} \in \{20, 30, 40, 50\}$ and no compression ($\tilde{n} = n$), and for the competitors.}
\label{tab:runtimes}
\begin{tabular}{lccccc|ccc}
\toprule
& $\tilde{n}=20$ & $\tilde{n}=30$ & $\tilde{n}=40$ & $\tilde{n}=50$ & $\tilde{n}=n$ & ACC & COBRAS & NCOBRAS \\
\midrule
Moons & 30.0 & 61.4 & 100.3 & 152.2 & 1030.2 & 12.1 & 0.4 & 37.3 \\
Iris & 29.6 & 61.2 & 99.3 & 151.1 & 2308.5 & 12.1 & 0.4 & 33.0 \\
Wine & 29.2 & 58.1 & 97.8 & 149.3 & 3233.3 & 12.0 & 0.5 & 38.0 \\
\bottomrule
\end{tabular}
\end{table}

% discussion

BACON has the best average rank at every query budget and noise level (\Cref{fig:ranking}). With noise BACON's average rank improves as the number of queries exceeds $q > 10$. This is because COBRAS degrades under noise and the BACON-rand falls behind due to less informative queries. In \Cref{tab:final}, we also report the best ARI attainable with the super-instance compression strategy. On 13 out of 16 datasets, BACON reaches within $0.05$ ARI of the bound. Notably, the super-instances for Glass and Waveform are quite impure, which explains the poor performance on these data.

With random queries, BACON often outperforms ACC, COBRAS and NCOBRAS. This suggests that much of the performance gain comes from the model itself. BACON-rand performed similarly to BACON on datasets with $k \leq 3$, so even random querying may suffice when the number of clusters is small But with $k \geq  5$ it fell clearly behind (see \Cref{tab:final}).
ACC performed the worst with no noise and improved the ranking as the noise level grew. It generally requires many more queries to reach the same performance as BACON, as it disregards the features vectors (used only in the initialization).

\Cref{fig:infering} summarizes the posterior samples of $\epsilon$ and $k$ on MNIST. For the first $30$ queries, $\epsilon$ varies around its prior mean of $0.15$, and after starts moving towards the empirical error rate $0.19$ with the variance decreasing slowly. The cluster count initially varies around $8$ and after some queries the posterior start to concentrate around $k_{gt}=10$.

{\bf Ablations.}
We studied BACON without compression ($\tilde{n}=n$) on data with limited number of samples (Moons, Iris, and Wine); for it, we set $\kappa = 10$ to ensure connectivity. The results are shown in \Cref{fig:abl_compression}. With a small number of queries ($q < 30$), compressed BACON performs better ($\tilde{n} \in \{20, 30, 40, 50\}$). With enough queries, BACON\textsubscript{$\tilde{n}=n$} can outperform them, attaining the best average rank after $50$ queries. The downside of compression is that the super-instances can be impure, which can limit the top attainable ARI (see \Cref{tab:final} column Bound). Without compression, BACON scales to hundreds of points, and with compression it can handle much more (we tested up to $n=70\, 000$). For instance, on Moons ($n=100$) the runtime per query for compressed BACON using $\tilde{n} = 50$ was 1.5 seconds, while BACON\textsubscript{$\tilde{n}=n$} took 10.3 seconds (see \Cref{tab:runtimes}).

We also evaluated BACON with different numbers of nearest neighbors $\kappa$. Using $\kappa \geq 5$ produced connected nearest neighbor graphs for all datasets using $\tilde{n}=50$. Using $\kappa=2$ produces disconnected nearest neighbor graphs on many datasets (e.g., MNIST), meaning that BACON cannot find the ground-truth clustering $\mathbf{c}^*$ if some $(i,j)$ with $\delta_{ij}(\mathbf{c}^*)=1$ resides in disjoint components of the nearest neighbor graph. The results are shown in \Cref{fig:abl_knn}. Overall, $\kappa=5$ achieves the best ranking, while $\kappa=2$ falls to last place after about $60$ queries because of disconnectedness. Using $\kappa \in \{10, 20\}$ produces worse results, but with enough queries they can perform similarly to $\kappa=5$. With small $\kappa$, there is less uncertainty to remove due to fewer links (see \Cref{sec:query_selection}), which is why we suggest setting $\kappa$ to a small value such that the nearest neighbor graph is connected.

Lastly, we studied whether inferring $\epsilon$ benefits the model by comparing BACON to variants with fixed $\epsilon$ to the correct noise level $0.1$, lower value $0.01$ and higher value $0.3$. From \Cref{fig:abl_noise}, we can see that using a wrong value is detrimental to the performance. Using a too low value can be better initially ($q < 10$), but the overconfidence costs later. Meanwhile, using a too large value costs early as the variant required more queries to reach similar performance. The cost of inferring the oracle noise level $\epsilon$ is insignificant, as the variant fixed to the exact value $0.1$ has very similar ranking across queries.

\section{Conclusion}

We proposed \emph{Bayesian Active Clustering with Oracle Noise} (BACON), a probabilistic active clustering method that combines a nearest-neighbor distance-dependent Chinese restaurant process with a noisy-oracle likelihood. BACON infers both the number of clusters and the oracle's error rate, and selects queries by expected information gain, which reduces to uncertainty sampling on posterior predictive must-link probabilities.
On 16 datasets with noise levels up to $\epsilon=0.2$, BACON achieved the best average rank at all query budgets. Even with random queries, BACON often outperformed the competing methods, which suggests that much of the performance gain comes from the probabilistic model itself.  Information-based selection nevertheless gave clearly better results on datasets with more than three clusters.

BACON has some limitations. The noise model assumes pair-independent errors (see \Cref{sec:query_selection}). Compressing the data into super-instances makes it scalable for large datasets, but errors in the initial over-clustering cannot be corrected. We resample the posterior after every query, which makes BACON slower than the heuristic methods COBRAS and NCOBRAS. A batched query selection could be used to speed up the algorithm, however, querying the oracle for the pairwise constraints remains the largest bottleneck.
Future work includes a split step for impure super-instances and a non-myopic selection criterion that would look more than one step ahead.

\begin{credits}

\subsubsection{\discintname}

The authors have no competing interests to declare that are
relevant to the content of this article.
\end{credits}

\bibliographystyle{splncs04}
\bibliography{bibliography}

\end{document}